\documentclass[pdflatex,sn-mathphys-num]{sn-jnl}

\usepackage{graphicx}%
\usepackage{multirow}%
\usepackage{amsmath,amssymb,amsfonts}%
\usepackage{amsthm}%
\usepackage{mathrsfs}%
\usepackage[title]{appendix}%
\usepackage{xcolor}%
\usepackage{textcomp}%
\usepackage{manyfoot}%
\usepackage{booktabs}%
\usepackage{algorithm}%
\usepackage{algorithmicx}%
\usepackage{algpseudocode}%
\usepackage{listings}%
\usepackage{caption}
\usepackage{multirow}

\theoremstyle{thmstyleone}%
\theoremstyle{thmstyletwo}%

\theoremstyle{thmstylethree}%

\begin{document}

\title[Hybrid Reasoning for Perception, Explanation, and Autonomous Action in  Manufacturing]{Hybrid Reasoning for Perception, Explanation, and Autonomous Action in  Manufacturing}


\author*[1]{\fnm{Christos} \sur{Margadji}}\email{cm2161@cam.ac.uk}

\author*[1]{\fnm{Sebastian} \sur{Pattinson}}\email{swp29@cam.ac.uk}

\affil[1]{\orgdiv{Department of Engineering}, \orgname{University of Cambridge}, \orgaddress{\street{\par Trumpington Street}, \city{Cambridge}, \postcode{CB2 1PZ}, \country{United Kingdom}}}


\abstract{Industrial processes must be robust and adaptable, as environments and tasks are often unpredictable, while operational errors remain costly and difficult to detect. AI-based control systems offer a path forward, yet typically depend on supervised learning with extensive labelled datasets, which limits their ability to generalize across variable and data-scarce industrial settings. Foundation models could enable broader reasoning and knowledge integration, but rarely deliver the quantitative precision demanded by engineering applications. Here, we introduce \textbf{C}ontrol and \textbf{I}nterpretation of \textbf{P}roduction via \textbf{H}ybrid \textbf{E}xpertise and \textbf{R}easoning (CIPHER): a vision-language-action (VLA) model framework aiming to replicate human-like reasoning for industrial control, instantiated in a commercial-grade 3D printer. It integrates a process expert, a regression model enabling quantitative characterization of system states required for engineering tasks. CIPHER also incorporates retrieval-augmented generation to access external expert knowledge and support physics-informed, chain-of-thought reasoning. This hybrid architecture exhibits strong generalization to out-of-distribution tasks. It interprets visual or textual inputs from process monitoring, explains its decisions, and autonomously generates precise machine instructions, without requiring explicit annotations. CIPHER thus lays the foundations for autonomous systems that act with precision, reason with context, and communicate decisions transparently, supporting safe and trusted deployment in industrial settings.}

\keywords{physical intelligence, embodied AI, manufacturing, hybrid reasoning, industrial AI}



\maketitle

\section{Introduction}\label{sec1}

\begin{figure}[t]
    \centering
    \includegraphics[width=1\linewidth]{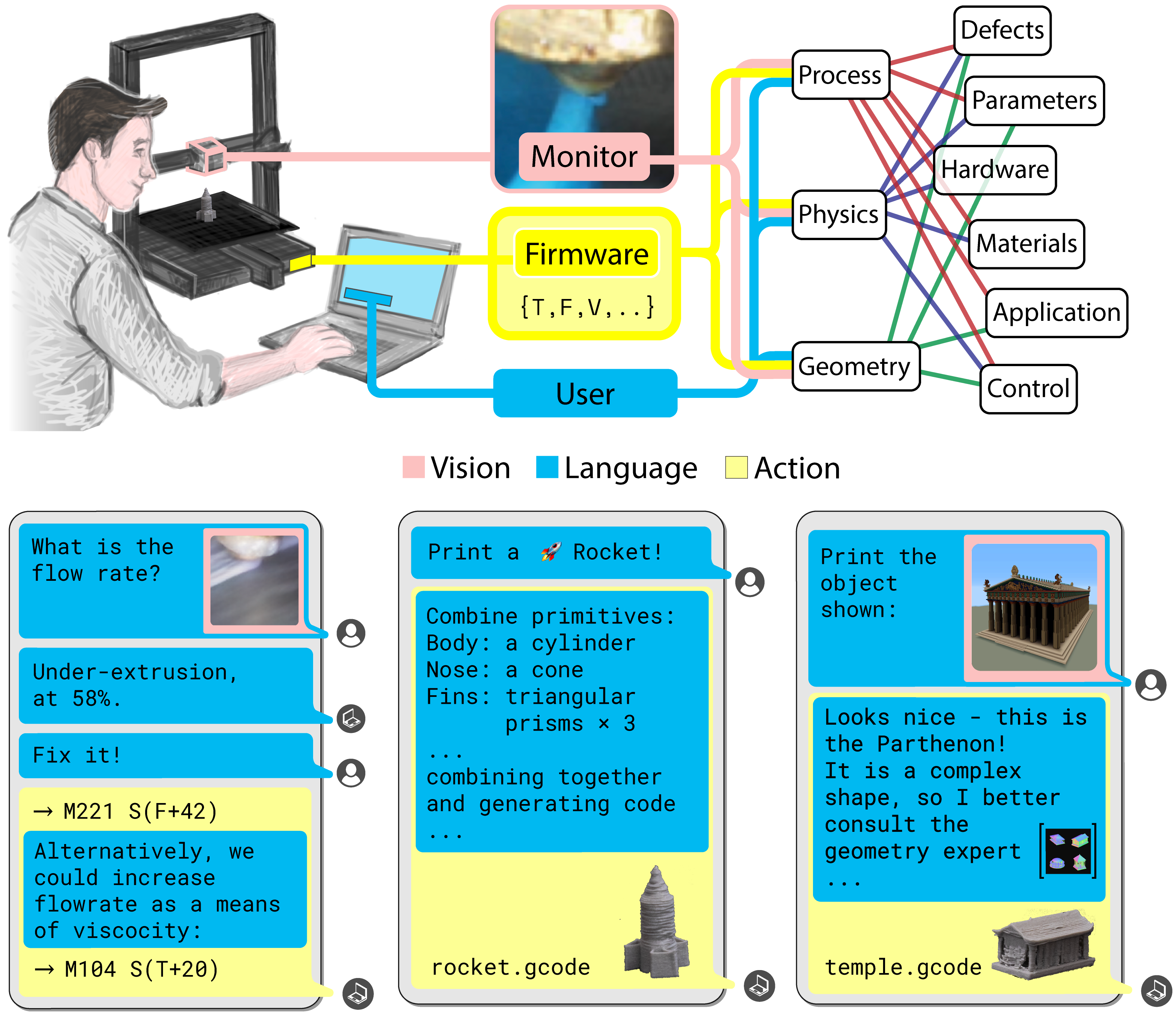}
    \caption{\textbf{Overview of the CIPHER framework.} Perception, reasoning and action are combined for robust and creative control in manufacturing.}
    \label{fig:introduction}
\end{figure}

Embodied agents perceive and act through a physical or virtual body, enabling interaction with and learning from their environment through sequential and synchronous sensory processing and decision-making. They typically integrate capabilities like language processing \cite{Achiam2023, Touvron2023}, visual perception \cite{Liu2023}, and action execution \cite{Zitkovich2023}, enabling decisions that are easily set and interpretable. Powered by their large foundational models, embodied agents exhibit remarkable emergent capabilities, such as creative problem-solving and adaptability \cite{Reed2022}, owing to training on diverse, web-scale data.\cite{Zhu2024, Muennighoff2024} It has been shown such agents can autonomously explore and interact with several environments, iteratively improving their own capabilities through self-verification, in turn out-performing gradient-based optimization reinforcement learning strategies.\cite{Wang2023}

Embodiment of intelligence in open-ended environments has demonstrated considerable promise, particularly in virtual (e.g., navigating web browsers and game worlds like Minecraft) but also physical settings.\cite{Zhou2023-web, Wen2024, Wang2023} In developing embodied agents for applications like autonomous driving, there has been significant progress, owning to the extensive availability of computational resources and training data, justified by the large-scale and homogeneous nature of the problem.\cite{Zhou2023-driving} Notable progress has been made in developing general physical intelligence, which can handle multiple tasks (particularly pick-and-place or other forms of object manipulation) with little to no retraining even for unseen hardware configurations.\cite{Wen2024, Kim2025} Still, commercial deployment of such intelligent systems remains limited owning to safety concerns as well as gaps relating to the Moravec's paradox (which states tasks being easy for humans are hard for robots and vice versa), these must be addressed.\cite{Zhou2023-driving, natMI_on_RobotLearning}

Training similar large-scale agents for industrial manufacturing robots lacks sufficient economic and environmental justification due to several constraining factors. One, the inherent diversity and smaller scale of manufacturing tasks, which largely vary on a per-case basis, leads to data inhomogeneity. Two, there are feasibility barriers when it comes to data collection and storage, mainly due to stringent confidentiality measures and reluctance to disclose manufacturing data (particularly from operations involving sensitive or intellectually protected assets.) Three, even when relevant data is available and accessible, the scarcity of cross-disciplinary expertise to combine manufacturing knowledge with that of machine learning is scarce, particularly in factory environments. We also add four, it is this lack of expertise that usually contributes to data quality degradation as a result of non-systematic collection pipelines, poor data engineering, and more.

Manufacturing tasks themselves exacerbate these challenges. High-dimensional, heterogeneous state spaces result in combinatorial complexity, while multi-physics interactions introduce dynamic, nonlinear, and often stochastic outcomes. The common absence of in-situ verification mechanisms does not help, permitting errors to propagate unchecked until post-process quality assurance occurs, thereby making it hard to establish error causality. Additionally, obtaining direct one-to-one correspondence datasets in manufacturing is often impractical, which further complicates causal inference through reliance on indirect or correlated observations. Quality metrics are also inherently quantitative when standardized, typically operating in continuous spaces. However, models at the frontier of physical intelligence, particularly large language models (LLM) and by extension vision-language (VLM) ones, exhibit weaker regression capabilities, require substantially more data compared to traditional deep learning approaches and need careful selection of loss functions before achieving comparable performance with traditional controllers.\cite{Song2024, kim2025finetuningvisionlanguageactionmodelsoptimizing}

Developing resource-efficient agents specifically tailored for embodiment in industrial equipment is critical. Here we adopt the hybrid reasoning paradigm, and rely less on large-scale models, instead leveraging smaller architectures for perception and utilizing larger ones primarily for reasoning and planning.\cite{Azarafza_2024, GUNTHER201290} This approach shapes our Control and Interpretation in Production via Hybrid Expertise and Reasoning (CIPHER) framework, a manufacturing agent explicitly designed to be easily trained and adequately robust for industrial deployment without sacrificing generalisation. CIPHER is exemplified here through embodiment within a commercial-grade 3D printing system, offering a realistic environment for automated control, sensing, and decision-making in machinery. Widely adopted across sectors such as aerospace, medical devices, and education, 3D printing excels at producing complex, customized parts.\cite{Klippstein2017,Zhang2022,Pattinson2017,Pattinson2019,Ford2019} Its multi-physics nature though, encompassing thermal, mechanical, and material interactions even in the most basic settings, introduces complexities.\cite{Zhang2021, Baechle-Clayton2022} These interactions require close monitoring - often necessitating expert human oversight - to prevent instabilities from developing into defects that compromise the functional integrity and usability of manufactured components.

Notable prior work on deep learning for quality assurance and control of 3D printing processes exists, and primarily focused on defect detection with more limited applications in correction. Sensor data, such as acoustic, inertial, pressure and current measurements, have been used to detect anomalies like nozzle clogs or filament jams.\cite{Wu2016, Shevchik2018, Pierre2023, Tlegenov2019} Vision-based sensors have also been used for more challenging problems such as detecting deformations in parts.\cite{Brion2022-warp} Recent advancements leveraged multi-headed deep residual networks for real-time defect and correction, tackling more complex scenarios like iterative improvements in mass production and discovering optimal process parameters for new geometries and materials.\cite{Brion2022-natcomms, Brion2022, Margadji2024} Additionally, it deep regularised neural fields have optimized process parameters even before printing to maximize geometric fidelity.\cite{Margadji2025} Nevertheless, although these methods excel at their own niche domains, none can address problems that have not been encountered during training, largely owing to their lack of causality and reasoning.

Conversely, our CIPHER framework integrates reasoning, vision-based perception, and natural language all under the same hood, to holistically overcome challenges related to robustness, data scarcity, and complexity that is inherent to industrial tasks. Central to our contribution is a process-expert integrated within a vision-language-action model, the process-expert enhances performance on the regression-oriented tasks that quantitatively characterize the state of the system. By effectively combining convolutional regression models' computational efficiency and reliability with foundational models' reasoning capabilities, CIPHER bridges the gap between precise engineering-grade perception and flexible, adaptive, explainable control. We demonstrate several interesting features, including emergent capabilities, which become better with retrieval-augmented generation (RAG) and physics-informed, chain-of-thought reasoning that enables robust adaptation to novel scenarios. A systematic analysis of the articulated chain-of-thought confers decision explainability and engenders the level of trust requisite in high‑stakes environments. Interestingly, we show that CIPHER can perform completely out of distribution tasks, such end-to-end autonomous fabrication from natural language or image prompts, linking abstract intent to physical execution. The agent’s high-level schematics are illustrated in Fig.\ref{fig:introduction}. 

\section{Results}\label{sec2}

\subsection{Training the vision-language-action model for 3D printing perception}

We begin with dataset preparation. A commercial endoscope is mounted on the head of the printer, capturing nozzle images as material is being deposited. These visual data streams are synchronized with corresponding real-time process parameters recorded from printer's firmware, which are hereby treated as our labels. Given the absence of linguistic annotations within these datasets, and the impracticality of manual annotation at scale, we systematically transform the collected numerical labels into structured natural language descriptions using predefined templates, followed by paraphrastic augmentation to increase linguistic diversity, as described in Methods and illustrated in Extended Data Fig. \ref{extended:fig2}. Every sentence contains a general statement about the process, followed by two statements that place the numeric parameter in meaningful context. With these image-caption pairs, the objective centers on the capability to accurately reconstruct the synthesized sentences (or semantically equivalent alternatives) solely from visual input, thereby and most importantly to retrieve the continuous values that have been embedded within.

We first benchmark a suite of publicly available multi-modal foundation models on this dataset. Preliminary evaluations demonstrate that our agent, without any fine-tuning and relying exclusively on prompt engineering, effectively addresses queries pertaining to general manufacturing contexts. Nonetheless, limitations emerge when the model is confronted with macroscopic observations or queries demanding detailed, granular insights into specific processes, Extended Data Fig. \ref{extended:fig1}. The observed discrepancy between general and granular-level tasks stems from the absence of manufacturing data in the original training corpus.

\begin{figure}[t]
    \centering
    \includegraphics[width=1\linewidth]{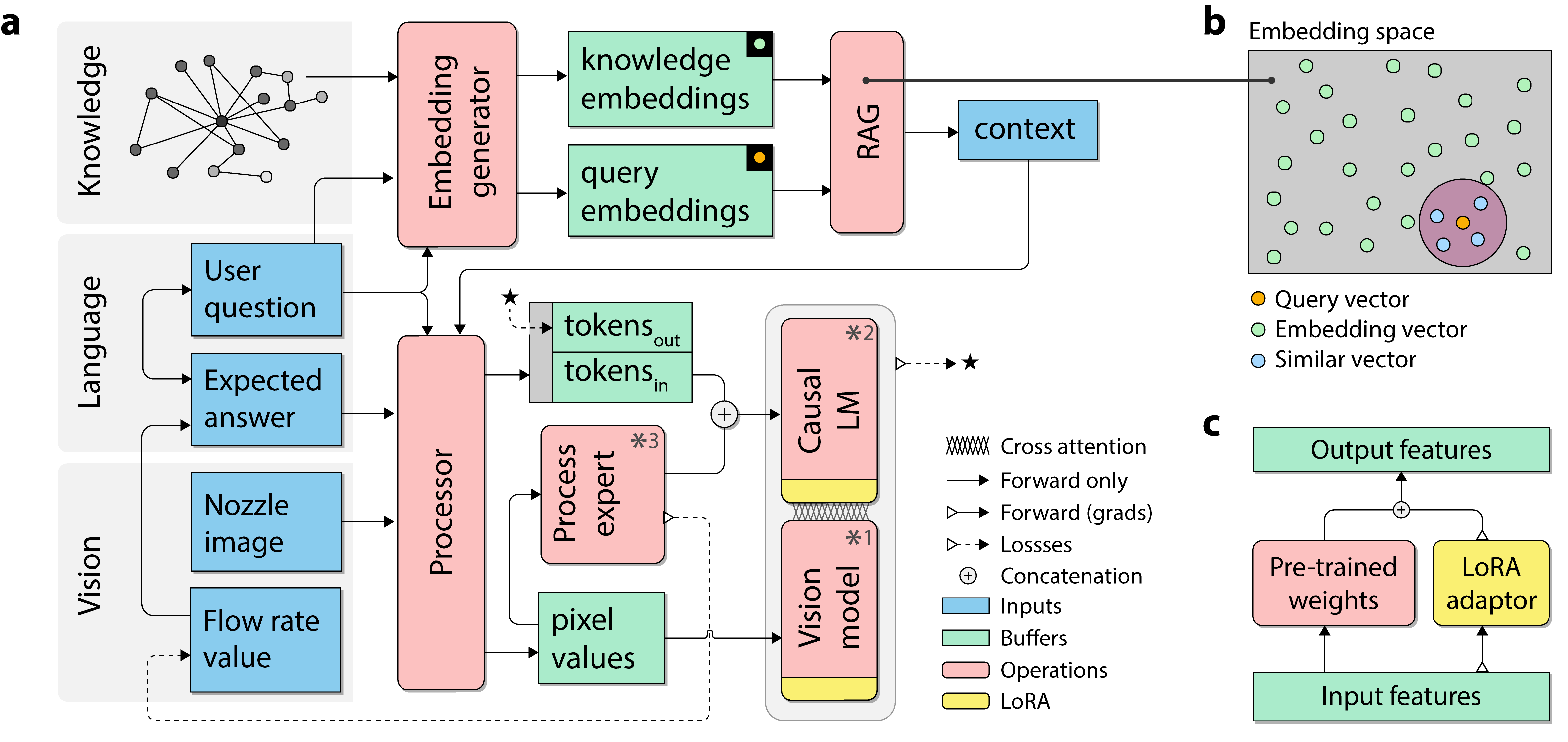}
    \caption{\textbf{The agent architecture.}
(a) The VLM architecture integrates a process expert which enables good qualitative and quantitative alignment. Details of the architecture can be found in Methods. (b) Illustration of the embedding space generated during retrieval-augmented generation. The query vector (orange) represents the input, while embedding vectors (green) populate the space. Similar vectors (blue) are identified within a defined proximity (purple region) to the query, enabling the retrieval of relevant information. (c) Schematic representation of the low-rank adaptation filters. The adaptor is introduced in parallel to the pre-trained weights of the architecture, which remain frozen, while the adaptor itself is fine-tuned.
}
    \label{fig:architecture}
\end{figure}

To minimise this discrepancy, we develop and train an architecture comprising a large language model (LLM) and a vision transformer (ViT), integrated via cross-attention layers as per the multi-modal Llama-3.2 framework, from which we also adopt pre-trained weights.\cite{Grattafiori2024} Within this architecture, we embed a convolutional neural network with residual connections between blocks, specifically utilizing a ResNet model, tasked with encoding visual inputs into task-relevant features.\cite{He2015} During inference, these features are linearly projected into token space and injected as a dedicated  start token that conditions the autoregressive decoder for captioning. During training, we feed the model the full sequence - expert, visual and textual tokens - and apply a causal  attention mask so that each position only attends to itself and earlier positions. We minimise the negative log‑likelihood of the ground‑truth sequence, optimising the cross‑entropy loss. All components - including the vision encoder, projection layer, and language backbone are trained fully end‑to‑end, more details in Methods and Fig. \ref{fig:architecture}a summarises the data flow.

\begin{figure}[t]
    \centering
    \includegraphics[width=1\linewidth]{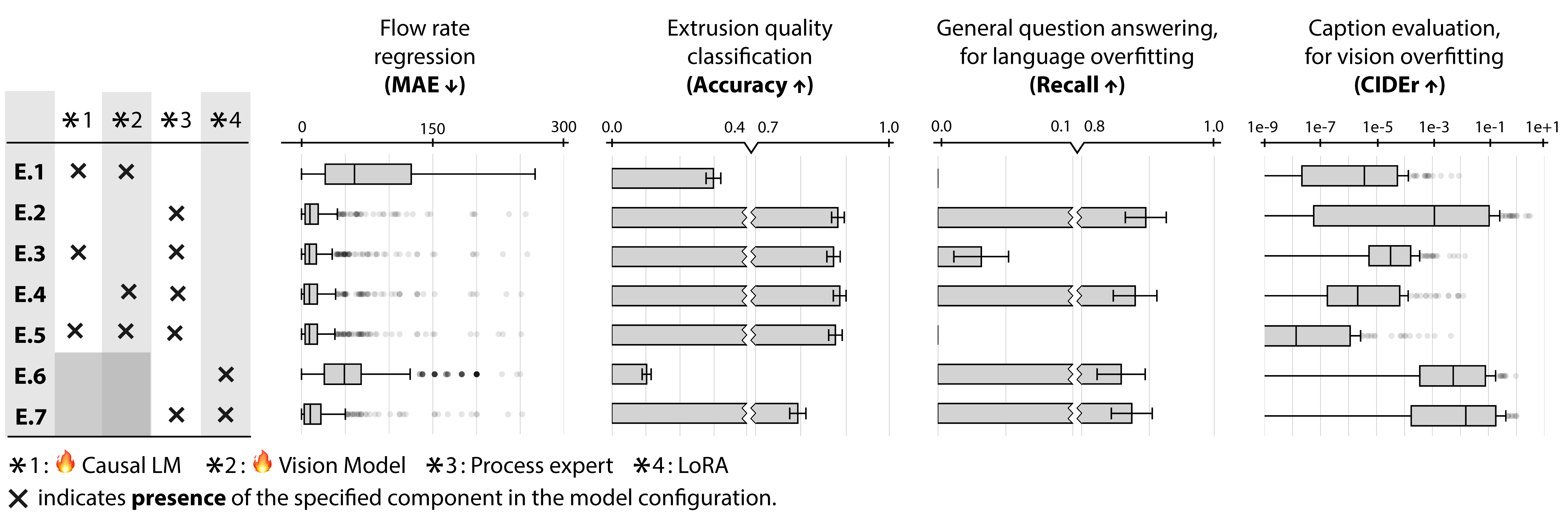}
    \caption{\textbf{Results from ablation studies.}
We conduct experiments on variants of our architecture to evaluate qualitative and quantitative alignment, as well as the model's susceptibility to catastrophic forgetting in both vision and language tasks.
}
    \label{fig:results}
\end{figure}

Post training we evaluate the model’s perception, focusing on quantitative and qualitative alignment with the domain. For qualitative alignment, we use classification accuracy to compare predicted classes against target classes mentioned in the generated sentences. For quantitative alignment, we extract from the text the continuous predictions and measure the discrepancy between predicted and ground‐truth values using mean absolute error (MAE) as in Methods. We also assess vision and language overfitting as an indication of catastrophic forgetting. All reported metrics are calculated from 1,000 held‐out test samples. Results from the various experiments conducted for this section are shown in Fig. \ref{fig:results}, with supplementary analyses provided in Extended Data Table \ref{extended:tab1}.

In the naïve fine-tuning experiments without the process expert, achieving language format and general domain alignment is trivial. Though, quantitative and qualitative information is not learned. Even when the whole network is trained for indefinite number of iterations, the mean absolute error (MAE) in the continuous flowrate predictions does not decrease, underscoring a key limitation in current vision-language models (VLMs) for accurately performing regression-based tasks. Similarly, the classification accuracy does not show any signs of convergence. This limitation is attributed to the classification-based loss function employed during gradient-based optimization. The non-deterministic nature of next-token prediction in the generation pipeline exacerbates the issue, and the problem persists even when the temperature is decreased to zero.

\begin{figure}
    \centering
    \includegraphics[width=1\linewidth]{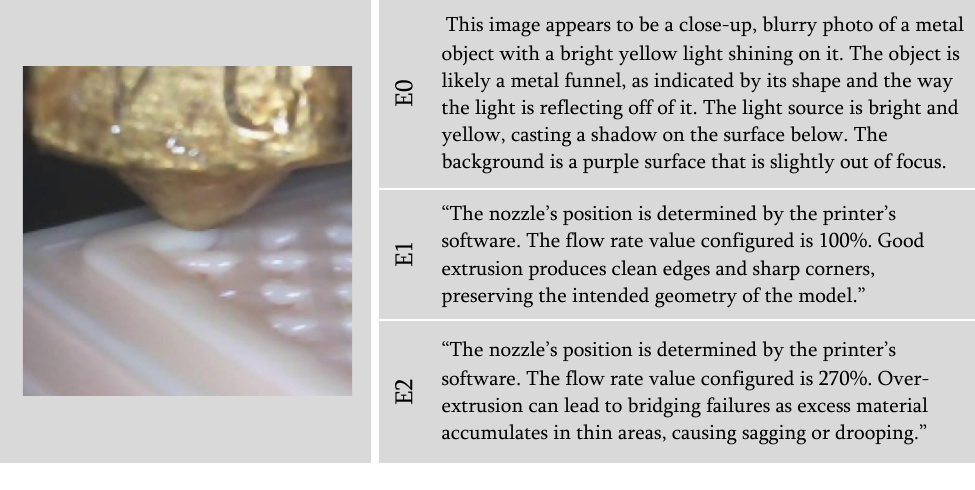}
    \caption{\textbf{Captions generated for the shown image from three models:} pre-trained (E0), fine-tuned without (E1) and with (E2) process expert. The ground truth flow rate value is 300\%.}
    \label{fig:data}
\end{figure}

To address this challenge, we adopt ideas from the meta-learning literature and integrate a convolutional neural network (e.g., a ResNet-152 (116m parameters) model) within our VLA. We refer to this model as the process expert, designed to perceive and retrieve task-specific information from the visual input. The process expert takes pixel-level information from the processor and delivers vision-specific insights directly to the language module in a compact, single-token format. This integration significantly improves the model's performance in regression tasks without adding substantial computational overhead. Quantitative and qualitative alignment show substantial improvement, with a five-fold reduction in MAE when comparing results from E1 (=82.92±71.83) and E5 (=17.62±30.04). The process expert effectively overcomes limitations inherent to the base architecture across all tested settings, even when the vision and language modules are frozen and only a small projection layer is optimized, that is in E2. In all observed cases, the generated qualitative descriptors consistently align with the flowrate predictions in a said sentence. Sample answers from the pre-trained model (E0), the trained model without (E1) and the trained model with (E2) the process expert are shown in Fig. \ref{fig:data}.

\subsection{Parameter-efficient fine-tuning}
We further investigate parameter-efficient fine-tuning strategies by employing Low-Rank Adaptation (LoRA).\cite{Hu2021} LoRA operates by freezing the core model weights and introducing auxiliary learnable matrices - termed adaptors - as depicted in Fig. \ref{fig:architecture}c. These adaptors are selectively optimized during fine-tuning to align the pretrained model with downstream task requirements. This methodology yields substantial efficiency gains, including a 52.4\% reduction in memory consumption (from 199.3 GB to 94.9 GB) and an 81\% reduction in the volume of training data required to reach comparable convergence. These gains are attributable to the drastically reduced number of trainable parameters, which constitute only 2.4\% of the model's total parameter count. Importantly, LoRA is applied uniformly across the end-to-end architecture; both the vision and language modules either include or exclude the adaptors in each configuration.

Quantitative and qualitative improvements are consistently observed only when the process expert is activated. While LoRA demonstrates strong compatibility with the process expert (E7), performance deteriorates by 7\% and 9\% on quantitative and qualitative metrics, respectively, when compared to the other strongest baseline (E5). This degradation is likely attributable to reduced network plasticity induced by the parameter constraints of LoRA - an effect we later show may be beneficial in mitigating overfitting and preserving generalization.

\subsection{Prior knowledge is preserved}

Retaining prior knowledge is critical to ensure reliable and robust reasoning that is not present in the training data. This makes the prevention of catastrophic forgetting of key importance. To evaluate the extend by which this has occurred or not, we compared the pre- and post-fine-tuning performance of the models on common natural language processing tasks, using 100 randomly sampled items from the SQuAD dataset for question answering and Flickr30k dataset  for image captioning.\cite{Rajpurkar2016,Plummer2016} Performance on these tasks is commonly validated through the language recall and CIDEr metrics as introduced in Methods.

Baseline performance metrics exhibit values of 0.965±0.015 and 0.412±0.433 for Recall and CIDEr, serving as reference points for subsequent evaluations. In experimental configurations E1 and E5, where joint fine-tuning of both language and vision modules is performed, we observe pronounced catastrophic forgetting. This manifests as a near-complete degradation in the two scores, suggesting that the model, even when presented with out-of-distribution inputs like images from the Flickr30k dataset, generates captions that are overly similar to the fine-tuning dataset suggesting overfit. While catastrophic forgetting persists when only one of the modalities is fine-tuned (in the absence or LoRA), its severity is notably attenuated. The application of LoRA substantially ameliorates overfitting effects, preserving 93.2\% and 95.6\% of the original language recall and CIDEr values, respectively, relative to the pre-fine-tuning baseline. Among all configurations, E2 emerges as the most robust against catastrophic forgetting. In this experiment, only the terminal projection layer of the vision module is adapted, leaving the core vision feature extractor and language generation parameters entirely intact. This selective adaptation strategy not only minimizes computational overhead but also yields superior qualitative and quantitative alignment with the target domain. Accordingly, E2 is designated as the preferred configuration for all subsequent analyses, owing to its efficiency, cross-modal generalization capacity, and immunity to catastrophic forgetting.

\subsection{Embodying CIPHER for action generation in novel settings}

\subsubsection{Performing traditional control}

In prior evaluations, configuration E2 demonstrated a high degree of accuracy in predicting flow rate values and articulating them in coherent, contextually grounded natural language outputs. Importantly, the model preserved its pretrained language generation capabilities, exhibiting no signs of degradation due to fine-tuning. Recognizing that programming languages (e.g., G-code, as defined by the ISO 6983 standard) constitute a machine-interpretable form of language, we extend our assessment to evaluate CIPHER's proficiency in generating executable commands. This task leverages the model’s dual competencies in linguistic reasoning and mathematical computation. Specifically, we provide the model with Supplementary Material Prompt S1, instructing it to perform the requisite numerical calculations and produce a structured M221 S{N} command, where {N} specifies the necessary flow rate adjustment. For context, the command M221 controls the system’s This experiment is designed to evaluate two core competencies: (a) the model’s ability to accurately translate conceptual input into structured machine code, and (b) its retention of arithmetic and symbolic reasoning capabilities following fine-tuning.

The experimental design proceeds as follows. We begin by uniformly sampling 100 distinct estimate values from the interval [30, 300], consistent with the empirical distribution of flow rate values observed in the original dataset. The term 'estimate' denotes predictions output by the process expert. Acknowledging potential discrepancies between these estimates and the internal state of the system - quantified as a mean absolute error of $17.52 \pm 28.89$ - we synthetically derive synthetic firmware values that represent the printer’s intrinsic measurement, or its belief, by sampling within a range defined by the said empirical error bounds, centered around the corresponding \verb|{estimate}|. This protocol captures the inherent uncertainty in model predictions and the variability in hardware-level belief formation. Each of the resulting 100 (estimate, firmware) pairs is supplied through Supplementary Material Prompt S1 as input to three models: our model, LLaMA 3.2, and GPT-4o-mini. All models are instructed to converge on a target flow rate of 100\%. Performance outcomes are presented in Fig. \ref{fig:emerging}a. CIPHER achieves parity with LLaMA and demonstrably outperforms GPT-4o-mini. With respect to the baseline MAE of $17.52$, the VLA introduces only a marginal additional control error of $0.215 \pm 0.090$, translating to an overall degradation of approximately $1.2\%$.

Latency performance remains strong, with belief estimation taking $1.5 \pm 0.8$ seconds while corrective adjustments are known to require another $0.8 \pm 0.4$ second. This results in a correction frequency of $2.3 \pm 1.2$ Hz. Even so, traditional control systems still respond faster. Using an intelligent agent just to imitate what these controllers already do well may not be necessary. More importantly, intelligent agents don’t overcome the key limitations of traditional controllers if they are also limited to a fixed set of actions, like discrete flow rate changes. This restriction reduces the agent’s ability to reason flexibly or explain its actions, which weakens its advantage over classical systems.

\subsubsection{Knowledge as context drives emerging behaviors}
\begin{figure}[t]
    \centering
    \includegraphics[width=1\linewidth]{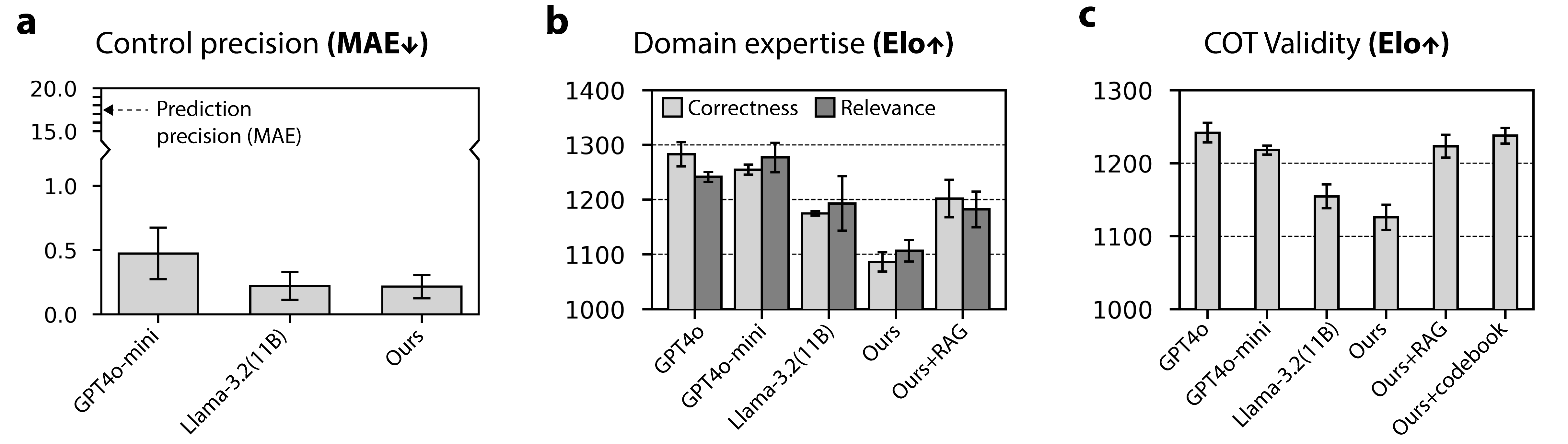}
    \caption{\textbf{Control performance of various agents across different scenarios.}
(a) MAE for in-distribution tasks (error bars: std, 1000 samples). (b) Elo ratings for QA tasks (correctness, relevance; std: 3 rounds, 100 questions each). (c) Elo ratings for chain-of-thought resoning for out-of-distribution tasks (std: 3 rounds, 100 challenges each).
}
    \label{fig:emerging}
\end{figure}
We have already discussed CIPHER's ability to perform hybrid reasoning, by combining observations with web-scale background knowledge, enabling it to operate within the domain even in unseen scenarios. This is particularly important in manufacturing but also engineering in general, in domains where one-to-one knowledge is hard to collect, and exploring or even simulating all possible scenarios may be extremely costly or risky. In this set of experiments, we test the agent’s knowledge on the domain, use RAG to enhance it, before using it as a proxy for emerging behaviors.

\textbf{Domain knowledge and knowledge injection.} To systematically assess the agent's domain-specific knowledge, we curated a set of 100 textbook-style questions derived from relevant literature, traversing the structured mind-map we present in Extended Data Fig.\ref{extended:fig3}. This mind-map was constructed with the assistance of an LLM to ensure comprehensive topic coverage. These questions were then posed to our model and a set of established baseline models, with comparative outcomes illustrated in Fig. \ref{fig:emerging}. To enhance performance evaluation, we also incorporated a retrieval-augmented generation (RAG) module, schematics in Fig. \ref{fig:architecture}b and detailed in Methods. This mechanism dynamically retrieves semantically aligned content from a pre-compiled knowledge repository, which itself was populated by prompting an LLM to generate factual statements for each node of the said mind-map.

To ensure statistically robust comparisons, we employed the Elo rating system - a metric typically utilized in competitive ranking scenarios (e.g. chess) - to evaluate the outputs in pairwise model matchups.\cite{Elo1978} This metric is commonly used in the LLM literature to compare performance between candidate models. In each trial, two models were randomly selected, and their respective answers were blindly evaluated by a third-party reviewer (implemented as an LLM), which judged responses based on correctness and contextual relevance. Each pairwise comparison was repeated across three independent rounds, with aggregated results summarized in Extended Data Table \ref{extended:tab2}. While minor evidence of catastrophic forgetting was observed in our model post-modification, the integration of the RAG module demonstrably mitigated these effects. More importantly, it conferred a net gain in factual accuracy and enabled the model to exceed its original performance baseline, achieving near-parity with state-of-the-art models in terms of relevance.

\textbf{Emerging behaviors.} Building upon our previous knowledge-injection experiments, we evaluate the agent’s emergent reasoning and control capabilities, particularly its capacity to generate corrective strategies in novel scenarios. To this end, we designed a benchmark comprising 100 randomized cases, each characterized by unique combinations of system settings (nozzle temperature, feed rate, z offset, working material), though we kept flow rate consistently fixed at 100\%. This explicitly incentivizes solutions that address other root causes rather than merely adjusting extrusion multipliers. Unique cases were encapsulated as $\verb|system_details|$
in Supplementary Material Prompt S4, and were provided to both our agent as well as other baseline models. Each model independently analyzed the inputs and produced structured outputs comprising: (a) a chain-of-thought rationale detailing its reasoning process, and (b) proposed control commands aimed at stabilizing or optimizing the system. This  framework allows a clear assessment of context-sensitive inference and adaptive control performance in unseen configurations, as well as a direct and fair comparison between models.

First we validate the chain-of-thought reasoning quality using the Elo ranking system like before (Supplementary Material Prompt 4), with results summarized in Figure \ref{fig:emerging}c and details in Extended Data Table \ref{extended:tab3}. Consistent with prior findings, our retrieval-augmented generation (RAG) model significantly outperforms the original Llama model, underscoring a positive correlation between robust question-answering capabilities and enhanced physics-informed reasoning. This confirms our hypothesis that domain-specific knowledge injection effectively reduces hallucinations and bolsters problem-solving in unfamiliar contexts. Crucially, providing the model access to the detailed G-code playbook (as described in Methods) and enabling explicit function calling markedly improved its performance (Elo$_{\text{Gcode}} = 1238\pm9$). This enhancement positioned our agent’s decision quality not only above GPT4o-mini (Elo$_{\text{GPTmini}} = 1218\pm5$) but also comparable to GPT-4o (Elo$_{\text{GPT4o}} = 1241\pm11$), highlighting the efficacy of explicit, structured domain knowledge integration.

CIPHER demonstrates adaptive, context-aware responses across novel control scenarios. It strategically adjusted nozzle temperature to modulate flowrate, leveraging the temperature-viscosity relationship, for instance, recommending, "Increase temperature to enhance melting efficiency and reduce flow resistance, as merely adjusting feed rates or firmware flow parameters would obscure underlying issues." It also accurately identifies mismatches between prescribed nozzle temperatures and material-specific melting points, such as recommending 240°C for ABS, and suggests corrective adjustments accordingly. When dealing with flexible filaments like TPU, the agent proactively reduces feed rate to prevent filament buckling. Additionally, the model occasionally suggests modulating fan speed to mitigate cooling artifacts, an unconventional yet valid strategy, and judiciously refrains from intervention when no corrections are required. Remarkable is one scenario wherein CIPHER proactively recommended aborting the print, recognizing the presence of unrecoverable faults. 

\subsection{Printing without explicit geometric supervision}

\begin{figure}[t]
    \centering
    \includegraphics[width=0.90\linewidth]{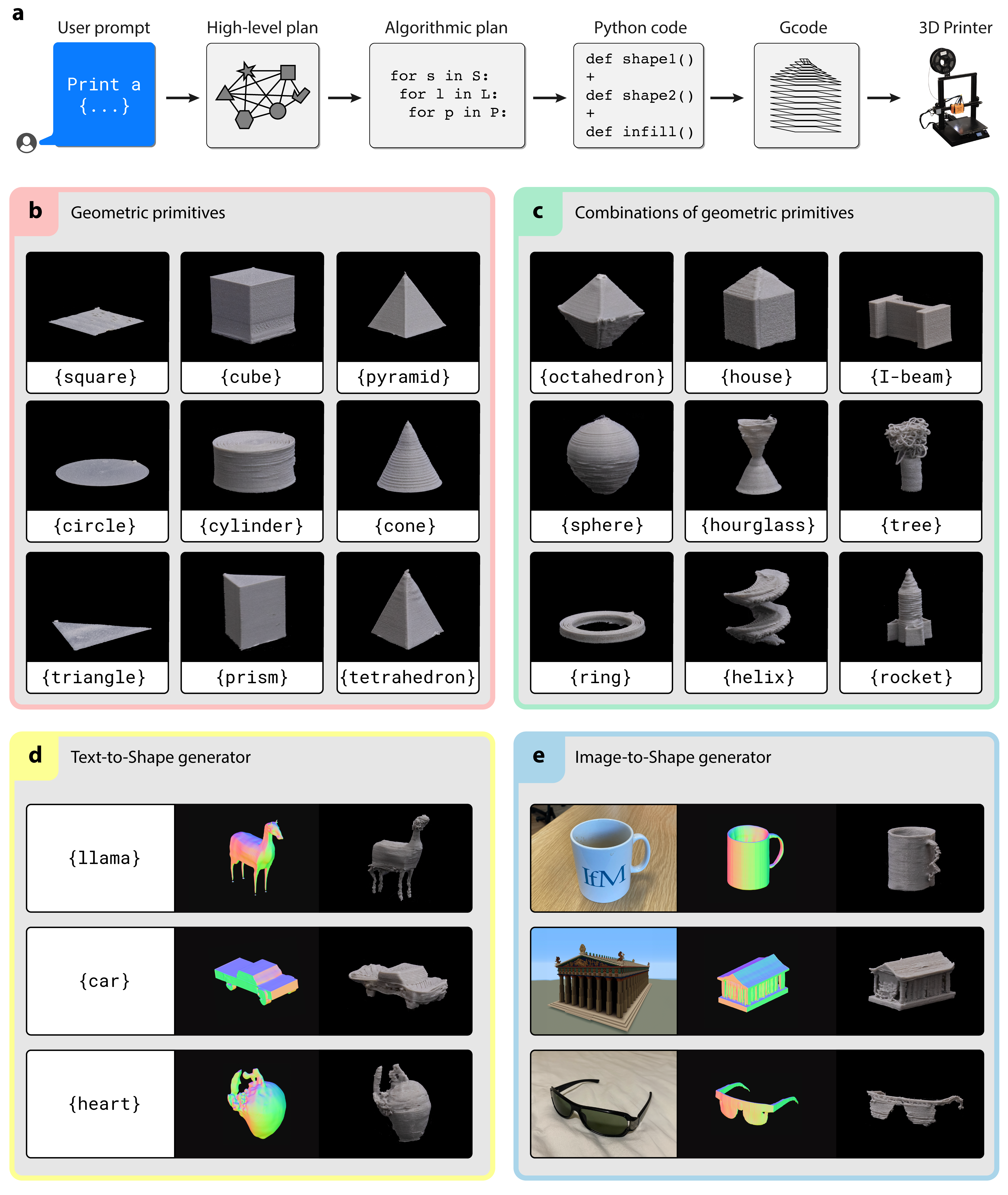}
    \caption{\textbf{Overview of translating abstract intent to physical execution through the geometry expert.} (a) The generation pipeline for geometric requests.  (b) Simple requests are represented by simple, well-defined geometric primitives. (c) Requests of medium complexity are represented as combinations of geometric primitives. (d,e) Complex requests are handled by the text/image to 3D shape generator.
}
    \label{fig:geometry}
\end{figure}

We finally test CIPHER on a completely out of distribution experiment to test its generalization. Here, we aim to show that reliance on foundation knowledge can even help with reducing reliance on human expertise in model preparation which critically impacts print quality. In our framework, we achieve this by embedding a 'geometry module', into the agent’s framework, enabling autonomous geometry inference and direct printing without manual preparation or oversight. We visualize the geometry module in Fig. \ref{fig:geometry}a, with low-level details provided in Methods.

Instead of relying on manual CAD design or mesh-based formats, we represent shapes as mathematical functions expressed in Python—either through compositions of geometric primitives or via implicit functions generated from natural language or visual inputs. These functional representations are subsequently translated into G-code instructions suitable for additive manufacturing, as illustrated in Fig. \ref{fig:geometry}b. While implicit function generation builds upon existing pretrained models (e.g., ShapE \cite{Jun2023}, our framework is the first to evaluate the direct primitive-composition route in a fully automated printing pipeline. We also introduce a novel filtering mechanism that dynamically selects between the two pathways based on the input type and structural complexity. The system enables fully automated, prompt-driven fabrication, as demonstrated in a range of examples—from simple commands like "print a cone" to moderate instructions such as "print a house", and complex generation tasks like "print a llama". Furthermore, we extend the pipeline to real-world deployment, enabling object printing directly from images captured in uncontrolled settings. All the geometries from our experiments are demonstrated in Fig. \ref{fig:geometry}b-e.

All geometric primitive requests succeeded in a one-shot manner. A key factor contributing to this success was starting with simpler 2D primitives, which were then incorporated into the "useful functions" section of Supplementary Material Prompt S5. These predefined functions provided efficient shortcuts and encouraged a consistent coding style, significantly simplifying the generation of more complex combinations. By leveraging this structured approach, the system could generalize effectively to more intricate 3D tasks, like building on the foundational primitives without redundant computation or increased error rates. For instance, a cone was synthesized by stacking progressively smaller circles on top of each other.

Synthesizing and printing complex 3D shapes by combining 3D primitives proved both challenging and feasible. Intricate forms like the helix (Fig. \ref{fig:geometry}c), which require precise offsets and rotations, succeeded in a one-shot manner. In contrast, simpler shapes such as the rocket, which involve only offsets, required iterative, few-shot optimization. While the agent effectively described objects as combinations of simpler primitives (e.g., a rocket with triangular fins, a cylindrical body, and a conical tip), it struggled with precise orientation and spatial alignment, particularly in correctly positioning details like the rocket’s fins. Additionally, the agent failed to account for the necessity of support structures (see Sphere), although it was capable of changing print orientation to minimize them (see I-Beam). Iterative adjustments to the prompts led to significant improvements, underscoring the potential for more spatially aware language models to handle these tasks more efficiently. Perhaps most exciting, in some cases the agent demonstrated creative solutions that exceeded human and traditional slicer capabilities, such as generating randomized extrusion paths to mimic the natural irregularity of the leaves (see tree.) 

Although any shape can theoretically be constructed by combining simpler primitives, this approach often becomes computationally intensive and impractical for complex designs. To overcome this limitation, our framework leveraged a pre-trained 3D shape generator capable of producing implicit 3D shape functions from text or image inputs. Novel to our approach was the way this was triggered, in cases in which more than N=4 primitives were needed for the original combination. In these experiments, the 3D shape generator produced some remarkable features, such as internal cavities resembling chambers (e.g. in the human heart), showcasing its potential for creating intricate designs. However, a significant challenge in this case was the frequent occurrence of unprintable features, arising because the 3D shape generator lacks an intrinsic understanding of manufacturing constraints. This manifests in structures that are visually plausible but mechanically fragile or unprintable due to the constraints inherent to fused filament fabrication, including violations of minimum printable feature dimensions (e.g., the legs of the llama) and geometries involving unsupported overhangs (e.g., the temples of the sunglasses), both of which can result in mechanical failure or necessitate auxiliary support structures during fabrication.

\section{Discussion}\label{sec12}

This work explores how hybrid reasoning in vision-language-action (VLA) systems can support autonomous control in data-limited, high-precision manufacturing environments. While large foundation models have demonstrated impressive generalization in language and vision tasks, their integration into physical domains has remained constrained by their limited ability to represent and reason over continuous, domain-specific parameters. CIPHER addresses this limitation through a modular architecture that combines a vision-language transformer with a dedicated regression expert and retrieval-augmented reasoning. The resulting system performs both perception and control, while maintaining interpretability and generalization, which are two properties typically at odds in industrial AI deployments.

The process expert plays a central role in enabling CIPHER to reason over continuous-valued physical states, such as material flow rate, which are critical in manufacturing but poorly represented in existing VLM architectures. This expert integration improves both quantitative prediction and semantic coherence between natural language and system state, suggesting a useful pathway for combining pretrained models with specialized modules when task precision is essential. Importantly, this approach circumvents the need for large annotated datasets, aligning with the reality of most manufacturing contexts, where labelled data are limited and costly to acquire.

The retrieval-augmented reasoning module allows the system to respond coherently to novel process conditions without direct supervision, leveraging external domain knowledge to generate physics-informed chain-of-thought explanations and actionable plans. This is a marked departure from prior data-driven controllers, which tend to generalize poorly outside training distributions. CIPHER’s capacity to produce interpretable rationales supports more transparent decision-making, although we realise the reliability of these explanations in safety-critical scenarios warrants further scrutiny.

A further area of investigation concerns the system’s spatial reasoning capabilities. While CIPHER is able to decompose geometric prompts into manufacturable primitives and generate corresponding G-code, its capacity to account for implicit constraints, such as material support, minimum feature size, or print orientation, remains limited. Instances of mechanical failure in complex geometries suggest that incorporating manufacturing-aware priors, either through constraint-based filtering or geometry-conditional training, will be necessary to improve robustness in autonomous fabrication tasks.

More broadly, this work raises questions about the role of modularity in embodied AI. CIPHER’s architecture suggests that targeted integration of process-expert and foundational components can offer a pragmatic alternative to end-to-end training when generalization and precision must coexist. At the same time, reliance on explicit module boundaries introduces interface challenges, for instance, ensuring semantic alignment between expert predictions and language model outputs, which merit closer attention in future research. In that way, CIPHER’s performance remains contingent on the quality of its expert modules and knowledge retrieval mechanisms, and further work is needed to evaluate the generality of this approach across domains, modalities, and fabrication processes. Nonetheless, these results suggest that foundation models, when appropriately constrained and supported, can move beyond abstract reasoning to support physically grounded, precise decision-making in real manufacturing environments.

\section{Methods}\label{sec11}

\subsection{Dataset}

\subsubsection{Vision}
The original dataset comprises more than a million nozzle images. In this work some samples are discarded due to being too blurry or overly like others, ensuring higher quality and diversity in the remaining data. During data loading, photometric and geometric augmentations are applied on a per-batch basis to increase robustness and generalizability. These augmentations included colour jittering (0-5\% brightness, contrast and hue shifts), rotations (-20° to 20°), translations (-10 to 10 pixels in both x and y axes), magnifications (80\% to 120\% of the original size), and random horizontal flipping (with p=0.5.)
\subsubsection{Language}
\textbf{Process perception.} In the dataset’s original format, each image is paired with a flowrate value retrieved from the firmware of the 3D printing system at the time of capture. To frame the problem as a vision-language task, we convert the flowrate value (our quantitative descriptor) into a natural language caption. Each of the captions includes:
\begin{itemize}
    \item A general statement about 3D printing.
    \item A quantitative sentence specifying the exact flowrate value.
    \item A qualitative descriptor based on the flowrate value: under-extrusion ($<$90\%), good extrusion ($\sim$100\%), or over-extrusion ($>$110\%).
\end{itemize}
For each caption component, a template is randomly selected from a pre-defined set, to prevent overfitting by diversifying. This maintains linguistic variability while ensuring accurate representation of the quantitative data.\\

\noindent
\textbf{Knowledge for retrieval-augmented generation}
We employ retrieval-augmented generation (RAG) for question answering, to enhance physics-intuitive chain-of-thought reasoning. To integrate domain knowledge into the pipeline, we first construct a comprehensive mind map of the 3D printing domain (Extended Data Fig. \ref{extended:fig4}). This mind map serves as a foundation for generating relevant facts. In total, 3,930 facts are generated, covering a diverse range of topics - from the chemistry of the materials involved to more specialized subjects, such as the history of the technology. These are converted and saved as embeddings using Ada-v2 provided from OpenAI's API, embedding space is plotted in 2D in Extended Data Fig. \ref{extended:fig5} using the t-SNE dimensionality reduction algorithm. We observe overlap between specific categories, such as physics and process parameters, offering intriguing insights into their interconnectedness. At inference time with RAG enabled, any prompt is encoded using the same embedding model and cosine similarity is used to find the N=5 nearest facts. These are then added within the prompt as additional context and passed to the agent for answering the original question.

\subsubsection{Action}
To facilitate accurate G-code generation, we parse the open-source Marlin firmware documentation (\url{https://marlinfw.org/meta/gcode/}) into a JSON format, where each key corresponds to a command name and its value includes a brief description and a URL (e.g., \url{https://marlinfw.org/docs/gcode/G000-G001.html}). During inference, the agent reasons over a task, converts the reasoning into an embedding using Ada-v2, retrieves the most relevant command using cosine similarity, and incorporates the fetched command's usage notes for generating precise G-code instructions.

\subsection{Agent architecture}

Our agent comprises of 3 main experts, namely process expert, physics expert and geometry expert. The process expert is responsible to perceive and understand the process, while the physics expert is responsible to turn observations into reasoning and actionable decisions. The geometry expert is responsible to handle geometric requests, allowing the agent to become more spatially aware.

\subsubsection{Vision Language module}

The architecture of the Llama-3.2-11B-Vision\footnote{https://huggingface.co/meta-llama/Llama-3.2-11B-Vision} is used, with the pre-trained weights loaded and used. The architecture comprises an image encoder, a projection layer, and a large language model (LLM). The image encoder, in our case a standard vision transformer ViT-H/14 \footnote{https://huggingface.co/laion/CLIP-ViT-H-14-laion2B-s32B-b79K}, transforms inputs from the pixel space into a high-dimensional feature space, while the projection layer maps these features into LLM-compatible tokens, enabling seamless integration with the language model for downstream processing and reasoning.

During training, the model is trained end-to-end using a next-token generation strategy, optimized with a cross-entropy loss function. At inference time for vision tasks, a processor prepares the image and attempts to reconstruct the reference caption, as defined in Methods. At inference time for language tasks, the input prompt is encoded, and the answer is generated in an auto-regressive fashion. Each token in the vocabulary is assigned a probability based on the model's predictions, and, following standard LLM practices, the next token is selected.\\

\noindent
\textbf{Parameter-efficient fine-tuning.}
We deploy low-rank adaptation (LoRA) for parameter efficient fine-tuning and hyperparameters are detailed in Extended Data Table \ref{extended:tab4}. While 4-bit quantization was explored to reduce the model's memory footprint, it significantly deteriorated the effectiveness of the process expert. This degradation is likely due to compatibility issues, as the vision components of the processor’s outputs serve as the inputs to the process expert. The reduced precision in 4-bit quantization introduces mismatches that disrupt this interaction, ultimately impacting the model's ability to maintain accurate and reliable performance.\\

\noindent
\textbf{Training.}
We train on a single node of 4 $\times$ NVIDIA A100-SXM-80GB GPUs, supported by 1TB of memory. Fine-tuning the model takes 20 hours over a single epoch, and a single epoch has been shown to be adequate for convergence. Hyperparameters and important network configuration settings are detailed in Extended Data Table \ref{extended:tab4}. Other details regarding training, inherently related to the selected architecture may be found in the original Llama paper and our codebase.\cite{Grattafiori2024}

\subsubsection{Process expert}
The process expert is a residual neural network ResNet-152\footnote{https://huggingface.co/microsoft/resnet-152} with 116 million parameters. Its primary role is to take as input a nozzle image of dimensions $224\times224\times3$ and explicitly predict the associated flow rate value. To adapt the architecture for this task, the final classification layer is replaced with a linear layer that maps 1024 features to a single output neuron. To fully leverage the pretrained weights, mean-std normalization is applied using the original statistics from the ImageNet dataset. Furthermore, the output space is transformed into logarithmic space, effectively mapping flow rate values from 30 to 300 to approximately -1 to 1.

During training, a batch of images is processed by the Llama processor and subsequently passed to the process expert. The expert generates predictions, and the ground truth values are used to perform an optimization step on the process expert. An AdamW optimizer is used and a linear learning rate scheduler ($\gamma$=0.5) is implemented with original value =1e-4. The updated predictions are then re-evaluated and converted from floating-point values to tokens, which are returned as input to the LLM alongside the original vision and prompt tokens. This iterative process ensures tight integration between the process expert and the LLM, enabling improved downstream performance.

\subsubsection{Geometry expert}
\textbf{Filtering.} The system uses a language model (LLM) to filter incoming requests, identifying whether they pertain to geometric tasks. Prompts containing commands like "Print a ... " are recognized as geometric requests and routed to the geometry expert module. In contrast, prompts involving queries such as "What do you see?" or "Is there any error?" are delegated to other specialized modules within the agent. This filtering mechanism ensures that the geometry expert focuses exclusively on shape generation and manipulation tasks, improving efficiency and task specialization.\\

\noindent
\textbf{Complexity tests.} 
All geometry requests undergo a complexity test to determine which branch of the geometry expert will handle the task. The complexity test leverages GPT-4o’s reasoning capabilities to assess whether a shape can be constructed using a combination of geometric primitives. Tasks requiring more than four geometric primitives are considered too complex and are routed to the text-to-shape generation module. Similarly, any request involving printing from images is sent directly to the image-to-shape generation branch.\\

\noindent
\textbf{Shape Generation}
For simple shapes, involving geometric primitives or their combinations, requests are reformatted based on Supplementary Material Prompt S5 and processed by the geometry module. As successful geometric operations are performed, the corresponding functions that generated these successful operations are added to the "useful functions" section. This approach builds a growing library of reusable, validated functions, enabling more efficient handling of future tasks by reducing redundancy and ensuring consistency. The prompt, enhanced by the accumulated "useful functions," is submitted to the GPT-4o API, which generates Python code automatically formatted into a .py file. This file is executed to produce a G-code file tailored to the requested operation. Before being forwarded to the 3D printing system, the generated G-code undergoes a validation step using the Cura Engine to ensure its executability and compatibility with the 3D printer.

For complex text-to-shape and image-to-shape requests, we use Shap-E\footnote{https://huggingface.co/openai/shap-e}. Shap-E generates implicit neural representations of 3D shapes, representing objects as continuous fields that determine whether a point is inside, outside, or on the shape’s surface. These representations are rendered into discrete 3D meshes using the marching cubes algorithm. To enhance user control, the system samples four shape options for each request, allowing the user to review and select their preferred design. The chosen shape is then passed to the subsequent module for G-code generation and, finally, to the 3D printer for execution.

\subsection{Validation metrics}
\subsubsection{Mean absolute error}
To evaluate the quantitative accuracy of predictions, we calculated the mean absolute error (MAE) between the predicted flowrate values and the ground truth values obtained from the 3D printing firmware. The MAE is computed as:
\begin{equation}
\mathrm{MAE}=\frac{1}{n}\sum_{i=1}^{n}\left|\widehat{y_i}-y_i\right|
\end{equation}
where $y\_i$ represents the ground truth value, $\widehat{y_i}$ represents the predicted value, and n is the total number of samples. This metric provides an intuitive measure of prediction accuracy, as lower MAE values correspond to better agreement between predicted and actual flowrates.

\subsubsection{Language recall}
To evaluate the alignment between predicted and ground truth language responses, we compute recall as the fraction of tokens in the ground truth that are correctly predicted by the model. Tokenization is applied to both predictions and references, and recall is calculated as:
\begin{equation}
R=\widehat{T}_{correct}/|T|
\end{equation}
where ${\hat{T}}\_{correct}$ is the number of correctly predicted tokens and $\left|T\right|$ is the total number of tokens in the ground truth reference. If the ground truth contains multiple references, the highest recall among them is used. This metric emphasizes the model's ability to capture essential elements of the reference text.

\subsubsection{CIDEr score}
Consensus-based image description evaluation (CIDEr) was developed to evaluate generated image captions by measuring their similarity to human-written reference captions. In the context of text-generation tasks (e.g., describing the quality of extrusion), CIDEr provides a score indicating how closely a generated response aligns with reference descriptions. Higher CIDEr scores indicate closer agreement with the reference text. It can be calculated using equation ()
\begin{equation}
    CIDEr_n(c_i,S_i) = \dfrac{1}{m} \sum_{j} \dfrac{g^n(c_i) \cdot g^n(s_{ij})}{||g^n(c_i)||\space||g^n(s_{ij})||}    
\end{equation}
where $c_i$ is the generated caption, $S_i$ is the set of reference captions, $g^n(c_i)$ is the n-gram frequency vector of $c_i$ and $g^n\left(s_{ij}\right)$ is the n-gram frequency vector of the $j$-th reference text. $m$ is the total number of available reference captions (in our experiments $m$=5 as commonly used with the Flickr30k dataset) and $\parallel\bullet\parallel$ represents the norm of the vector, ensuring proper normalization.

\subsubsection{Qualitative-Quantitative alignment}
The qualitative-quantitative alignment has been designed to evaluate whether the predicted flow rate aligns with the described extrusion quality in the generated text in a classification fashion. Specifically, it checks if a flow rate below, equal to, or above 100\% is correctly described as under, good, or over-extrusion, respectively. Alignment can then be expressed as a ratio of correct predictions to the number of total predictions.

\subsubsection{Elo ranking}
The Elo ranking system, originally developed for chess, was employed to evaluate the relative performance of different agents in pairwise comparisons. Each agent’s rating was updated based on the outcomes of these comparisons, reflecting their relative skill levels. The Elo ranking formula is given by:
\begin{equation}
R^\prime=R+k\cdot (S-E)
\end{equation}
where $R^\prime$ is the updated Elo rating, R is the current rating, k is the update factor, S is the score achieved in the match, and E is the expected score, calculated as:
\begin{equation}
    E=\dfrac{1}{1+10\times\dfrac{R_{opp}-R}{400}}
\end{equation}
Here, $R_{opp}$ represents the opponent’s rating. For our experiments, we set k=16 and initialized all agent ratings at 1200. The score S is assigned as S=1 for a win, and S=0.5 for both agents in the case of a draw. To determine scores for a pair of answers, we consulted a GPT-4o instance using Supplementary Material Prompt S3 for question-answering and Supplementary Material Prompt S4 for physics-informed chain-of-thought reasoning.




\backmatter
\clearpage

\bmhead{Supplementary information}

We include prompts and accompanying code, including for experiments, at the project's repository: \url{https://github.com/cam-cambridge/CIPHER}.

\bmhead{Acknowledgements}

We would like to acknowledge funding from the UK Research and Innovation (UKRI) Interact Centre (J17293 / ES/W007231/1) and Engineering and Physical Sciences Research Council awards EP/V062123/1 and EP/N509620/1. 

\bmhead{Conflicts of Interest}

SWP is co-founder and CM is an intern and future employee of Matta Labs, a company that sells AI-based software for the manufacturing industry. All other authors declare no conflicts of interest.










\begin{appendices}
\clearpage
\section{Extended Data}\label{secA1}

\begin{table}[h]
\caption{\textbf{Ablation study for the process module.}
E.0: original Llama model with the pre-trained weights. $\star$1: trainable language module, $\star$2: trainable vision module, $\star$3: Process expert; $\star$4: low-rank adaptation enabled. 
}
\begin{tabular}{c|cccc|cccc}
\hline
    & $\star$1 & $\star$2 & $\star$3 & $\star$4 & \begin{tabular}[c]{@{}c@{}}Regression\\ MAE\end{tabular} & \begin{tabular}[c]{@{}c@{}}Classification\\ Accuracy\end{tabular} & \begin{tabular}[c]{@{}c@{}}Language\\ overfitting\end{tabular} & \begin{tabular}[c]{@{}c@{}}Vision\\ overfitting\end{tabular} \\ \hline
E.0 &   &   &   &   & n/a                                                      & n/a                                                               & 0.96 ± 0.02                                                    & 0.41 ± 0.43                                                  \\
E.1 & \checkmark & \checkmark &   &   & 82.92 ± 71.83                                            & 29.77 ± 2.03                                                      & 0.01 ± 0.01                                                    & 0.00 ± 0.00                                                  \\
E.2 &   &   & \checkmark &   & 17.52 ± 28.89                                            & 88.25 ± 1.46                                                      & 0.90 ± 0.03                                                    & 0.38 ± 0.49                                                  \\
E.3 & \checkmark &   & \checkmark &   & 17.30 ± 29.33                                            & 87.28 ± 1.48                                                      & 0.04 ± 0.03                                                    & 0.00 ± 0.00                                                  \\
E.4 &   & \checkmark & \checkmark &   & 17.62 ± 30.04                                            & 88.59 ± 1.47                                                      & 0.88 ± 0.03                                                    & 0.00 ± 0.00                                                  \\
E.5 & \checkmark & \checkmark & \checkmark &   & 17.78 ± 29.95                                            & 87.57 ± 1.42                                                      & 0.00 ± 0.00                                                    & 0.00 ± 0.00                                                  \\
E.6 &   &   &   & \checkmark & 61.95 ± 52.92                                            & 10.15 ± 1.41                                                      & 0.86 ± 0.04                                                    & 0.12 ± 0.13                                                  \\
E.7 &   &   & \checkmark & \checkmark & 19.07 ± 31.13                                            & 79.46 ± 1.79                                                      & 0.88 ± 0.03                                                    & 0.24 ± 0.21                                                  \\ \hline
\end{tabular}
\label{extended:tab1}
\end{table}

\begin{table}[h]
\caption{\textbf{Elo ratings for question answering.}
We conducted three rounds of assessments, scoring each model’s responses based on three metrics, correctness, conciseness, and relevance. Higher Elo ratings indicate better relative response quality.
}
\begin{tabular}{cccccc}
\hline
\multicolumn{1}{c|}{}        & Ours & Ours+ RGA & Llama-3.2 & GPT-4o mini & GPT-4o \\ \hline
\multicolumn{6}{c}{Correctness}                                                    \\ \hline
\multicolumn{1}{c|}{Round 1} & 1093 & 1197      & 1174      & 1253        & 1283   \\
\multicolumn{1}{c|}{Round 2} & 1066 & 1238      & 1171      & 1264        & 1261   \\
\multicolumn{1}{c|}{Round 3} & 1099 & 1171      & 1179      & 1246        & 1305   \\ \hline
\multicolumn{6}{c}{Relevance}                                                      \\ \hline
\multicolumn{1}{c|}{Round 1} & 1099 & 1219      & 1143      & 1308        & 1232   \\
\multicolumn{1}{c|}{Round 2} & 1129 & 1170      & 1194      & 1266        & 1242   \\
\multicolumn{1}{c|}{Round 3} & 1092 & 1157      & 1243      & 1258        & 1251   \\ \hline
\end{tabular}
\label{extended:tab2}
\end{table}

\begin{table}[h]
\caption{\textbf{Elo ratings for physics-informed chain-of-thought reasoning.}
We conducted three rounds of assessments, scoring each model’s responses based on three metrics, correctness, conciseness, and relevance. Higher Elo ratings indicate better relative response quality.
}
\begin{tabular}{c|cccccc}
\hline
        & Ours & \begin{tabular}[c]{@{}c@{}}Ours + RGA\end{tabular} & \begin{tabular}[c]{@{}c@{}}Ours + codebook\end{tabular} & Llama-3.2 & \begin{tabular}[c]{@{}c@{}}GPT-4o mini\end{tabular} & GPT-4o \\ \hline
Round 1 & 1139 & 1220                                                 & 1232                                                      & 1141      & 1211                                                  & 1257   \\
Round 2 & 1131 & 1210                                                 & 1231                                                      & 1173      & 1224                                                  & 1231   \\
Round 3 & 1106 & 1240                                                 & 1250                                                      & 1150      & 1218                                                  & 1236   \\ \hline
\end{tabular}
\label{extended:tab3}
\end{table}

\begin{table}[h]
\caption{\textbf{Hyperparameters used to train our CIPHER framework.}}
\begin{tabular}{ll}
\hline
\multicolumn{1}{l|}{Parameter}                 & Value                                                              \\ \hline
\multicolumn{2}{l}{Global}                                                                                          \\ \hline
\multicolumn{1}{l|}{Num Train Epochs}          & 1                                                                  \\
\multicolumn{1}{l|}{Train Batch size}          & 4                                                                  \\
\multicolumn{1}{l|}{Gradient Steps}            & 8                                                                  \\
\multicolumn{1}{l|}{Learning rate (Llama)}     & 2e-4                                                               \\
\multicolumn{1}{l|}{Learning rate (v. expert)} & 1e-4                                                               \\
\multicolumn{1}{l|}{Max. gradient norm.}       & 0.3                                                                \\
\multicolumn{1}{l|}{Warmup ratio}              & 0.03                                                               \\
\multicolumn{1}{l|}{Max Length}                & 2014                                                               \\ \hline
\multicolumn{2}{l}{LoRA (applied only on VLA)}                                                                      \\ \hline
\multicolumn{1}{l|}{alpha}                     & 16                                                                 \\
\multicolumn{1}{l|}{dropout}                   & 0.05                                                               \\
\multicolumn{1}{l|}{Rank}                      & 8                                                                  \\
\multicolumn{1}{l|}{Target modules}            & \begin{tabular}[c]{@{}l@{}}queries,\\ keys,\\ kernels\end{tabular} \\
\multicolumn{1}{l|}{Quantization}              & True, 4-bit                                                        \\ \hline
\multicolumn{2}{l}{Expert}                                                                                          \\ \hline
\multicolumn{1}{l|}{Warmup batch size}         & 16                                                                 \\
\multicolumn{1}{l|}{Learning Rate}             & 1e-4                                                               \\
\multicolumn{1}{l|}{Scheduler}                 & Step                                                               \\
\multicolumn{1}{l|}{Warmup iterations}         & 500,000                                                            \\ \hline
\end{tabular}
\label{extended:tab4}
\end{table}

\begin{figure}[h]
    \centering
    \includegraphics[width=1\linewidth]{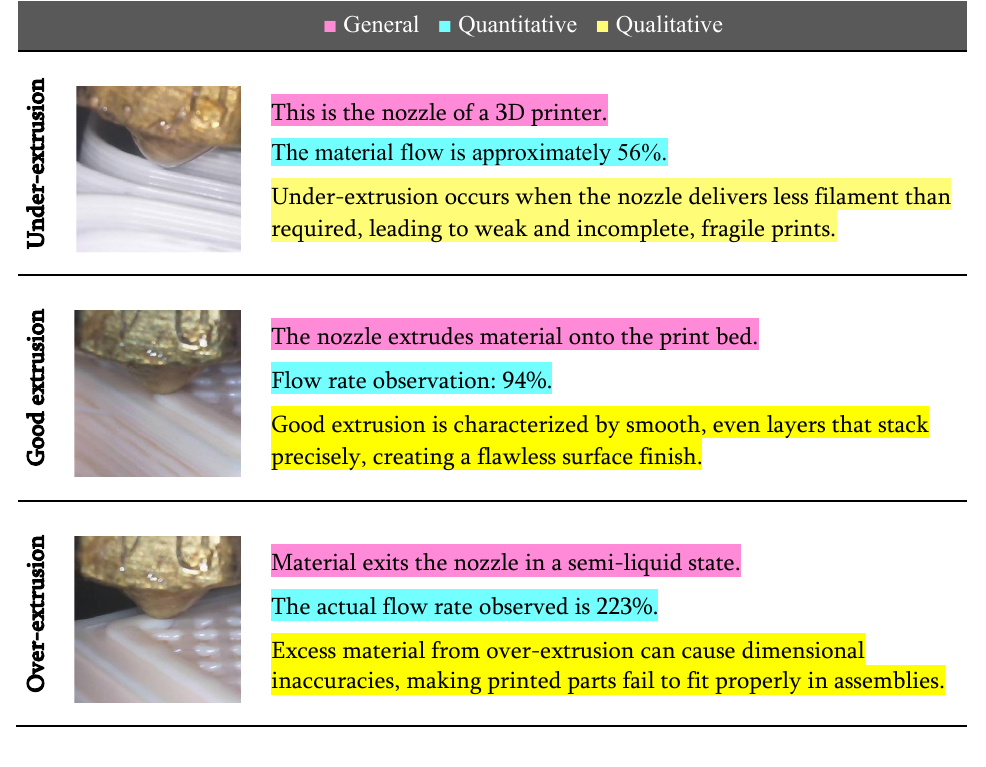}
    \caption{\textbf{Example image-caption pairs used to train the CIPHER.}
Captions are divided into three components: general, quantitative, and qualitative. Based on the continuous flow rate label, samples are categorized as over-extrusion, under-extrusion, or good extrusion.
}
\label{extended:fig2}
\end{figure}

\begin{figure}[h]
    \centering
    \includegraphics[width=1\linewidth]{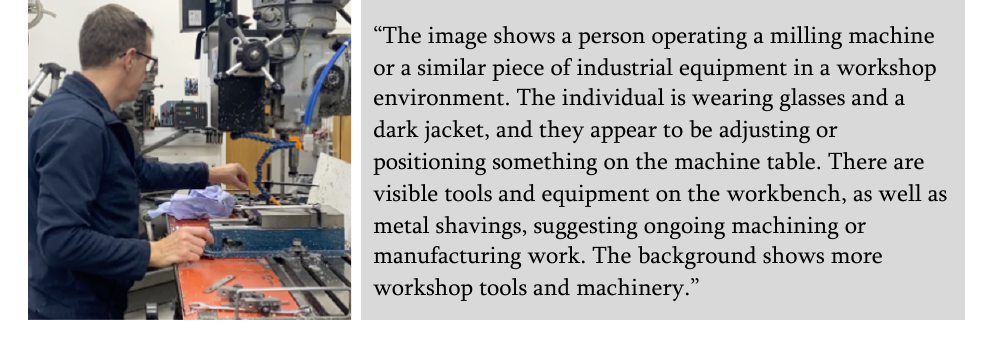}
    \includegraphics[width=0.96\linewidth]{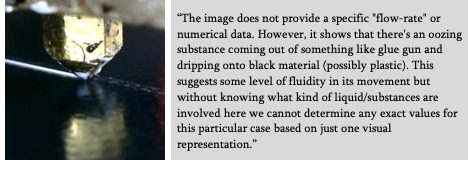}
    \caption{\textbf{Top:} Response on high-level manufacturing scenarios. \textbf{Bottom:} Response on low-level manufacturing scenarios in which macro inputs are given, and granular information is requested.}
    \label{extended:fig1}
\end{figure}

\begin{figure}[h]
    \centering
    \includegraphics[width=1\linewidth]{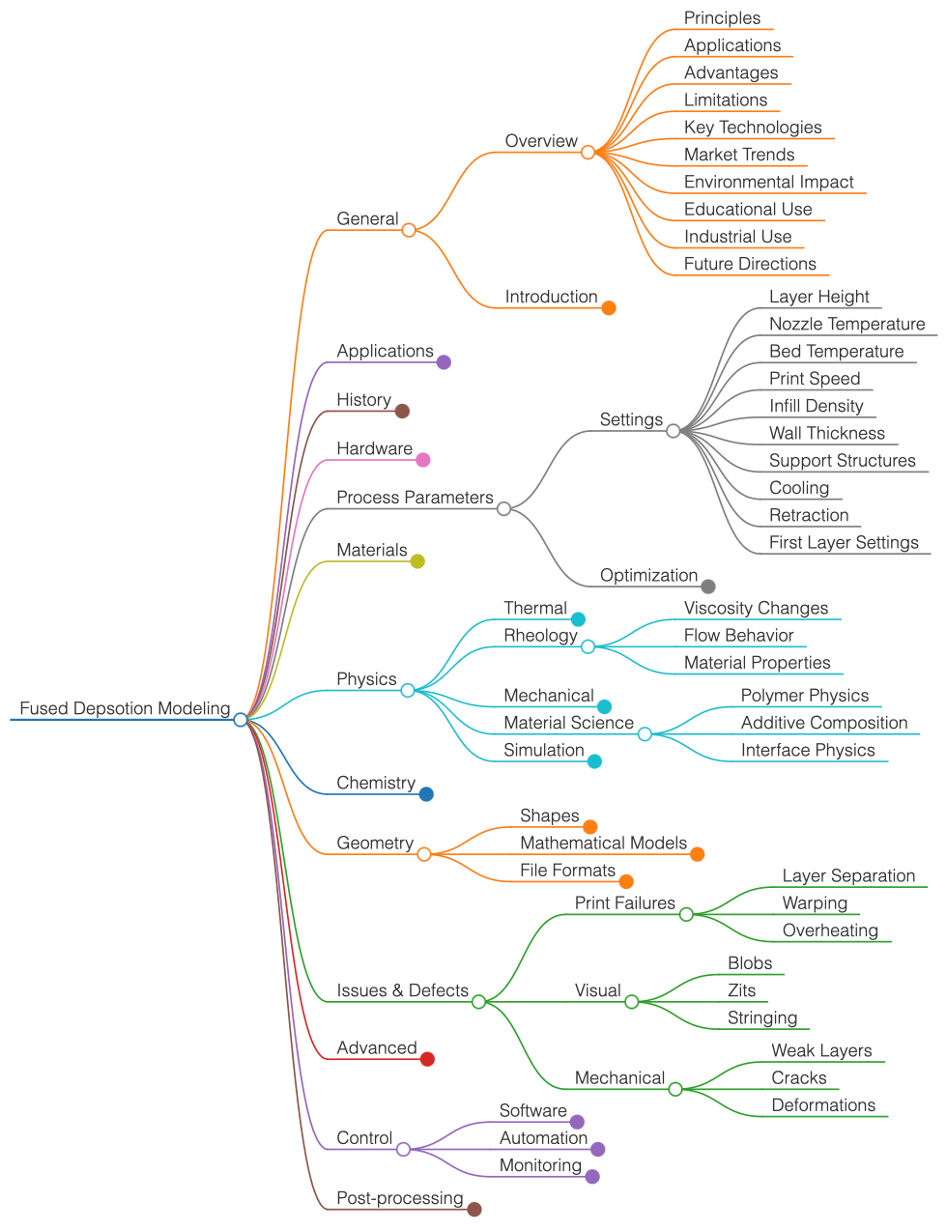}
    \caption{\textbf{Mind map used to guide the fact generation used for retrieval-augmented generation.}}
    \label{extended:fig3}
\end{figure}

\begin{figure}[h]
    \centering
    \includegraphics[width=1\linewidth]{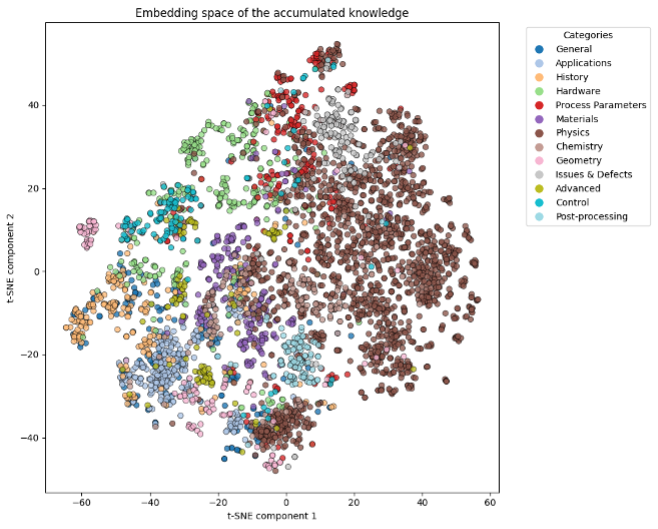}
    \caption{\textbf{T-SNE plot of the embedding space of all gathered knowledge used for retrieval-augmented generation.}}
    \label{extended:fig4}
\end{figure}

\begin{figure}[h]
    \centering
    \includegraphics[width=1\linewidth]{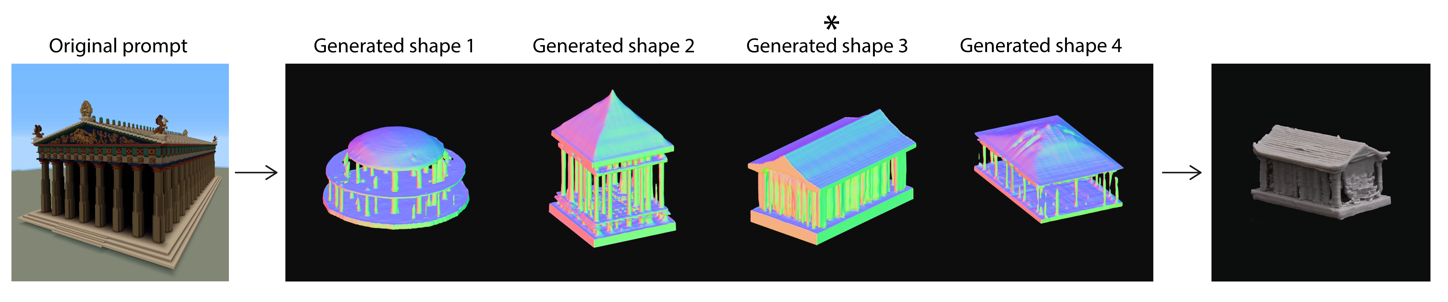}
    \includegraphics[width=1\linewidth]{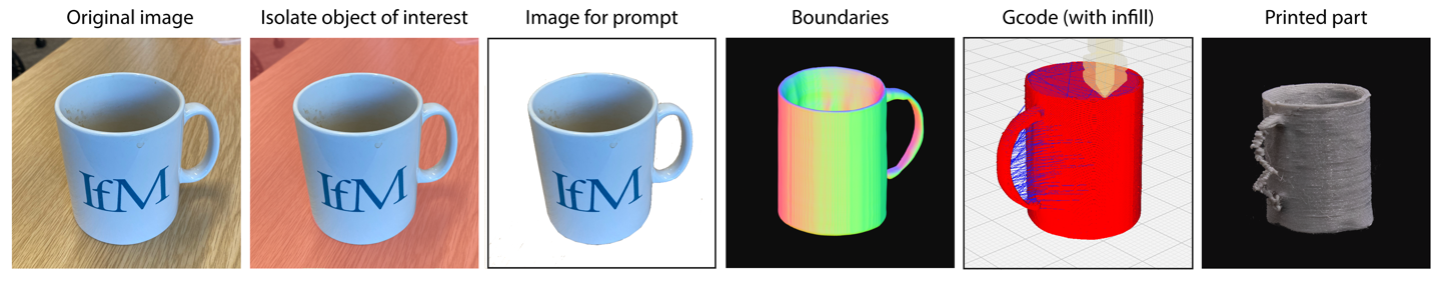}
    
    \caption{\textbf{Top:} The agent returns four shape as options for the user to select from, before converting to Gcode and initiating the printing process. * indicates the user's choice in this instance. \textbf{Bottom:} The end-to-end pipeline from image to 3D printed object. Some filtering and other pre-processing (e.g. segmentation with SAM\cite{kirillov2023segment}) is done by the geometry expert.}
    \label{extended:fig5}
\end{figure}




\end{appendices}


\clearpage
\bibliography{citations}

\end{document}